# An Indian Roads Dataset for Supported and Suspended Traffic Lights Detection

Sarita Gautam[1(*)],
Research Scholar, Panjab University, Chandigarh, India.
Email: Gautamsarry@gmail.com[(*)]

Anuj Kumar[2],
Assistant Professor, Panjab University, Chandigarh, India.
Email: Anuj_gupta@gmail.com

**Abstract**: Autonomous vehicles are growing rapidly, in well-developed nations like America, Europe, and China. Tech giants like Google, Tesla, Audi, BMW, and Mercedes are building highly efficient self-driving vehicles. However, the technology is still not mainstream for developing nations like India, Thailand, Africa, etc., In this paper, we present a thorough comparison of the existing datasets based on well-developed nations as well as Indian roads. We then developed a new dataset "Indian Roads Dataset" (IRD) having more than 8000 annotations extracted from 3000+ images shot using a 64 (megapixel) camera. All the annotations are manually labelled adhering to the strict rules of annotations. Real-time video sequences have been captured from two different cities in India namely New Delhi and Chandigarh during the day and night-light conditions. Our dataset exceeds previous Indian traffic light datasets in size, annotations, and variance. We prove the amelioration of our dataset by providing an extensive comparison with existing Indian datasets. Various dataset criteria like size, capturing device, a number of cities, and variations of traffic light orientations are considered. The dataset can be downloaded from here https://sites.google.com/view/ird-dataset/home

## 1. Introduction:

Vision-based systems are trending these days. They are gaining popularity for self-driving cars and Driver Assistant systems. In well-developed nations, self-driving cars are becoming popular due to their safety, efficiency, and convenience. On the contrary in a highly-populated country like India, where technical problems (non-working traffic signals, missing traffic signals) and non-technical problems (highly-congested roads, multiple intersections, trees, obstructions) still prevail on the roads, self-driving technology has a long way to go. It's challenging for self-driving technology to run efficiently on highly congested and non-maintained roads of underdeveloped nations. Talking about the challenges Traffic Light perception and detection is one of the major problems in accident-proof self-driving in India because the traffic lights are not well maintained and there might also be non-working traffic lights. Additionally, the challenges like recognition of moving objects such as cars, pedestrians, cyclists, motor-cyclist, and static objects like traffic lights and poles also hinder the efficiency of autonomous vehicles. These kinds of problems are being well addressed by researchers for well-developed nations. However, very few researchers have worked on Indian road data and a negligible amount of Indian data is available. Here we are proposing a new dataset "Indian Roads Dataset" (IRD) based on Indian roads captured from the roads of New Delhi and Chandigarh in day & night conditions. This dataset contains images from two different cities and presents traffic lights in two different orientations namely supported and suspended. Section II represents the state-of-art datasets based on well-developed and under-developed nations, section III shows the limitations and drawbacks of the existing datasets, section IV provides a detailed description of our IRD dataset, and section V represents the annotation tools that we have used for labelling and annotations. Section VI presents a technical evaluation of IRD dataset.

## 2. Review of Existing Datasets based on well-developed nations:

In Table-1 we present an overview of existing traffic light datasets based on well-developed nations. Table-1 clearly depicts that the datasets based on well-devolved nations are shot using high-definition cameras and have large number of high-quality images in them. The authors have classified various objects on the road by labelling different classes like car, pedestrians, pole, traffic lights, motorcycle, bicycle, truck, traffic cone, road pile, fence, traffic light, pole, traffic sign, building, bridge, tunnel, overpass and many more. A Tabular comparison of datasets based on well-developed nations is given in Table-1 and Table-2.

| Reference | Dataset name | Camera | Size | Location | Resolution | Conditions | Classes | Tested on |
|---|---|---|---|---|---|---|---|---|
| [1] | CDTD | Sekonix AR0231 | 155,529 | Tokyo, Japan | 1280 X720 | Sunny, night, cloudy, rainy | Car, person, traffic-sign | NA |
| [2] | Kitti vision Benchmark suit | Stereo camera | 12,919 | Karlsruhe | 1280 X720 | day | Car, Pedestrian | NA |
| [3] | Cityscapes | Stereo camera | 5000 fine annotations, 20000 coarse annotation | Various cities | 1024 X 2049 | Day light | Car, truck, bus, bridge, tunnel, Pole, traffic light, traffic sign, motorcycle, bicycle etc. | NA |
| [4] | SIM10k | GTA V engine | 10000 images of cars | NA | 1280 X720 | Clear, Overcast, Rainy, Night, Light snow | car | NA |
| [5] | nuScenes | 8 cameras have been used to capture images. | 93,000 2d annotated images | Boston, Pittsburgh, Las Vegas and Singapore | 1600x1200 | Clear, Overcast, Rainy, Night, Light snow, heavy snow | Car, Person, animal,stroller,wheelchair,barriers,debris,motorcycle, trailer, truck,police,ambulance, bendy, construction_worker etc. | NVIDia TITAN GPU |
| [6] | BDD100k | NA | 120,000,000 | New York, Berkely, san-francisco, | 1280 X720 | Clear, rainy, snowy, overcast, cloudy, foggy, average | Person, car, sign, light, truck, bus, bike, rider, motor, train | NA |

| | | | | bay-area | | | | |
|---|---|---|---|---|---|---|---|---|
| [7] | Apollo Scape | Stereo camera, HD cameras | 5000+ frames | NA | 3384 X 2710 | Instance-level, pixel-level, Depth images | Vehicles, pedestrians, riders, car, truck, traffic cone, road pile, fence, traffic light, pole, traffic sign, building, bridge, tunnel, overpass | NA |
| [8] | SJTU small traffic light dataset | NA | 5786 | NA | 1080 X 1920 | Flickering/fluctuating traffic lights, illumination/exposure changes, multiple visible lights, false positives | Red, yellow, green, off, wait | NA |
| [9] | DriveU traffic light dataset | Stereo camera | 232,039 | Germany | 640 X 480 | Daylight | Circle, arrow straight, arrow left, arrow straight left, arrow right, pedestrian, cyclist | NA |
| [10] | YBY | AVT Moke G-125c camera with 12-36na mm Lens | 30,886 | Urban areas of Beijing, china | 1292 X 964 | Daytime, Nighttime | Go, stop, go_ arrow, left_go, right_go, left_stop | 4 NVIDIA Jetson TX1 and two FPGs |
| [11] | NA | NA | 71,770 | San-francisco, California | 1280 X720 | Day, night, city, residential | Green, Red, Yellow, green(left-turn), red(left-turn), yellow(left-turn) | 3 NVIDIA Titan X GPUs |

Table 1: Extensive overview of Datasets based on Well-developed nations

All the existing datasets based on well-developed nations are mentioned in section-II. To provide a comparative analysis, we compared the characteristic value of Kitti, Nuscenes, BDD100k, Apollo Scape, and DriveU datasets. Their infographics are given in Figures 1-6.

1. **Kitti:** Kitti vision benchmark dataset provides real road images for autonomous driving. The dataset contains more than 12000 images having 1382 x 512 resolution shot on 10FPS and divided in 11 different classes namely buildings, sky, tree, cars, sign, road, fence, pole, sidewalks, pedestrians, bicyclist etc. The infographics of Kitti dataset is provided in Figure 1.

2. **NuScenes:** A large-scale dataset for autonomous driving captured in Boston and Singapore using 6 cameras. 32-beam LiDAR and radars with 360-degree view cameras having about 10 classes such as cars, trucks, buses, trailers, pedestrians, construction vehicles, motorcycles, bicycles, barriers, and traffic cones. This dataset contains 1,400,000 images leading it to be one of the largest datasets publicly available. The infographics of NuScenes dataset is provided in the Figure 2.

3. **BDD100k:** This is a diverse dataset for heterogenous learning for multiple tasks. BDD100k was shot in multiple cities such as Berkeley, New York, San Francisco and Bay Area. The dataset contains more than 100,000 images and have classes like bus, light, sign, person, bike, truck, motor, car, train and rider. The infographics of BDD100k dataset is provided in the Figure 3.

4. **Apollo Scape**: An autonomous driving dataset shot in multiple cities and sites having high variance. The dataset contains more than 100,000 images captured from 10 different cities. The infographic for Apollo Scape has been shown in Figure 4.

5. **DriveU:** This dataset has been generated to be one of the largest and high variance datasets, this dataset contains more than 23000 annotations divided into 344 classes and captured from 11 different cities. The infographic for DTLD has been shown in Figure 5.

| Dataset | Resolution | Frame Rate | Annotations/Images | Cities | Classes |
|---|---|---|---|---|---|
| Kitti | 1382 X 512 | 10 Fps | 12919 | 1 | 11 |
| nuScenes | 1600x1200 | 20Fps | 1,400,000 | 2 | 10 |
| BDD100k | 1280 x 720 | 30Fps | 100k | 4 | 10 |
| Apollo Scape | 3384 x 2710 | 30Fps | 140,000 | 10 | 30 |
| DriveU | 2048 X 1024 | 15 Fps | 232,039 | 11 | 344 |

Table 2: Characteristics comparison of different traffic light datasets.

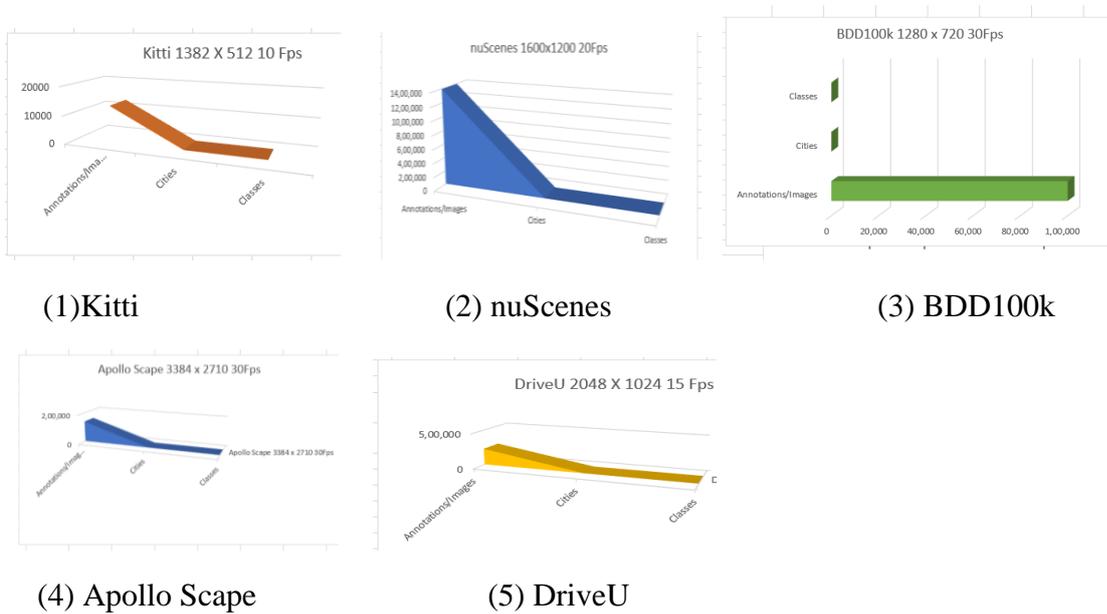

(1) Kitti     (2) nuScenes     (3) BDD100k

(4) Apollo Scape     (5) DriveU

**Figure (1-5)**: Frequency Comparison between five datasets based on well-developed nations. Each of the infographic here presents number of annotations/images, number of cities the dataset have been shot and the number of classes in every dataset.

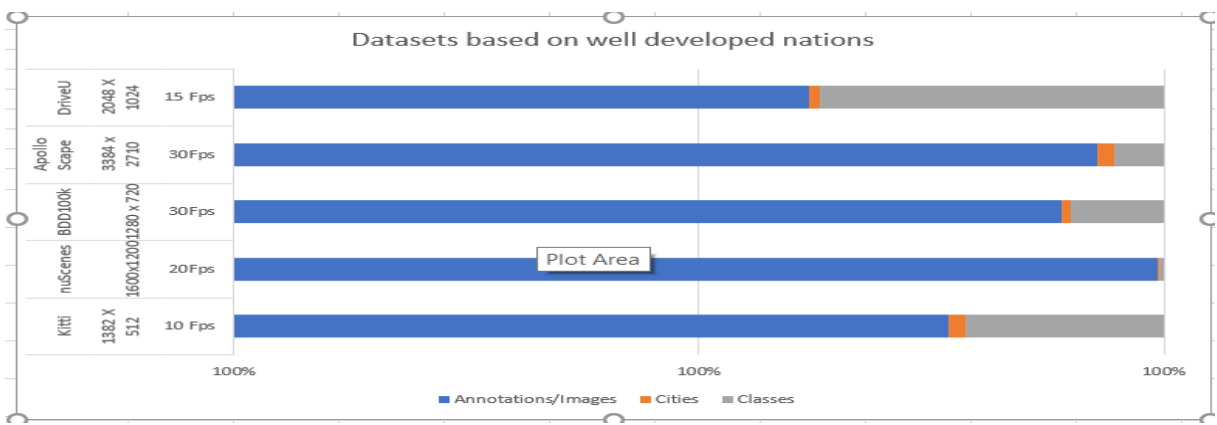

**Figure 6:** shows all the datasets based on well-developed nations in a comparable form.

## 2. Overview of Existing Datasets based on Indian roads:

Table-3 presents an overview of existing traffic light datasets based on Indian roads. After analysing Table 1 and Table-3, it can clearly be defined that the datasets based on Indian roads are not that mature. These datasets are captured using poor camera devices and have limited number of images in them. On top of that none of the datasets are made publicly available. Here we are providing our very own dataset captured from two Indian cities and making it publicly available to use. Further details are mentioned in section 5.

| Reference | Dataset name | Camera | Size/Annoations | Location | Resolution | Conditions | Classes | Tested on |
|---|---|---|---|---|---|---|---|---|
| Ours | IRD dataset | Redmi note-9 pro max, having 64 megapixel camera. | 3,334 (images + thumbnails) /8000+ annotations | New Delhi & Chandigarh, India | 3840 X 2160 | Daylight, night | Car, traffic-light, bus, person, Auto-rickshaw, bicycle, truck, motobike, green traffic light, red-light, timer, yellow light, green-light | NVIDIA GEFORCE GTX 1650 |
| [12] | Private dataset | 16 megapixel | 1237 | Pune, Maharastra | 4616 X 3464 for images and 1920 X 1080 | daylight | Straight, right, left, yellow, green | NVIDIA GEFORCE 940M GPU |
| [13] | Private dataset | Normal video camera | Live feed videos | Indian roads | NA | daylight | Car, motorcycle | 6th Generation i5 clocked at 3.0GHz with Catalina MacOS and 8 Gigabytes RAM |
| [14] | Private dataset | Normal video camera | 12 hours long video | Chandigarh, from PGI to Madhya Marg | NA | Day, night | Car, motocycle | Webster's method |
| [15] | Used Tesla's autopilot features, BMW's LiDar | Forward looking side camer, rearward looking side camera, rear view camers, radar sensor's data, | NA | NA | NA | NA | NA | Arduino uno, IR sensor |
| [16] | NA | Video footage of 50 frames per second | NA | Tamil nadu | 336x 596 | daylight | Cars | Matlab , Arduino circuit |
| [17] | Private dataset | Video frames and still images | NA | Bengluru | NA | daylight | Lane, potholes, road signs images | NA |
| [18] | Private dataset | Still images of road sign, roads with potholes, road images with lane markings | NA | Indian roads | NA | daylight | Lane, potholes, road signs images, human shapes, bounding boxes | MatLab |
| [19] | Used publicly available BSTLD | Images are acquired from a publicly available dataset | Still images | Not on Indian roads | 1280x720 | Busy streets, overcast, light rain, flickering light, illuminations, | Traffic light | NA |

| | [20] | Private dataset | NA | NA | Nashik , India | NA | NA | Traffic control management, traffic analyzer | NA |
|---|---|---|---|---|---|---|---|---|---|

**Table 3:** Extensive overview of Datasets based on Indian roads

All the existing datasets based on Indian roads are mentioned in section 3. To provide a comparative analysis, we compared the characteristic value of Our dataset with the other dataset. As analysed from table 3 not much information is provided about the Indian datasets. So, we present an infographic of these datasets in comparison to our IRD dataset. The infographic of Indian roads datasets is given in Figure 7. Since the Indian datasets are private so they do not have any name, here we are presenting these datasets by the author's name.

| Dataset | Resolution | Frame Rate | Annotations/Images | Cities | Classes |
|---|---|---|---|---|---|
| IRD | 3840 X 2160 | 3FPS | 8000+ | 2 | 14 |
| Kulkarni's | 4616 X 3464 for images and 1920 X 1080 | 20Fps | 1237 | 1 | 5 |
| Sharma's | Not provided | NA | NA | 1 | 2 |
| Sharma's | Not provided | NA | NA | 1 | 2 |
| Ritwik's | Not provided | NA | NA | 1 | 1 |
| Kumar's | Not provided | NA | NA | 1 | 3 |
| Danti's | Not provided | NA | NA | 1 | 5 |
| Swati's | Not provided | NA | NA | 1 | 2 |

**Table 4:** Characteristics comparison of different Indian traffic light datasets.

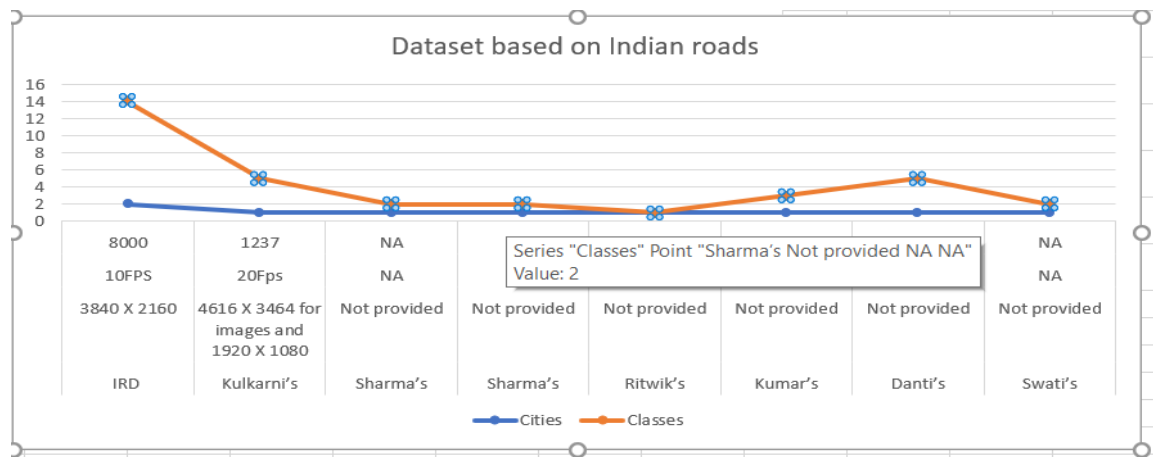

**Figure 7:** Infographic of dataset based on Indian roads.

**4. Limitations in Existing datasets**: As analysed from Table-1and Table-2, we can see that the dataset shot in well-developed nations/cities has a large number of images in them, captured using high-definitions cameras or multiple cameras. Their capturing vehicles are well equipped with high-definition cameras, LiDar, radar sensors, etc. These datasets are highly useable as compared to Indian datasets which have been shot using minimal quality devices. The Indian datasets are not well organized and also a limited number of objects have been considered in them. So here we propose a new dataset having 3000+ images (combined with thumbnails) captured from the roads of New Delhi and Chandigarh shot in a day and night-light conditions using a 64 mega-pixel camera having a total of 8,520 annotations of different objects. Our dataset provides a better insight into Indian roads based on two different cities namely New Delhi and Chandigarh. We have considered 13 classes for different objects in our dataset as mentioned Car, traffic light, bus, person, Auto-rickshaw, bicycle, truck, motorbike, green traffic light, red light, timer, yellow light, and green-light.

**5. Our Contribution (IRD):** We collected video sequences from the roads of New Delhi and Chandigarh cities of India. We used wagonR car and mounted a camera inside the car using a camera-holder. We used a mobile phone with a camera of 64 Megapixels for capturing the video sequences from two cities New Delhi and Chandigarh. In the city, Chandigarh video sequences from Housing board chowk to Panjab University were captured. It's a 10 k.m. long route as shown in map Figure-8. For the New-Delhi video sequence from Govind-puri to Gurgaon, it's a 10 k.m were captured. long route as shown in map Image-2. After capturing the videos, we extracted frames from them using the VLC media player with a frame of 20 fps. We have a total of 3050 images combined from both New Delhi and Chandigarh roads.

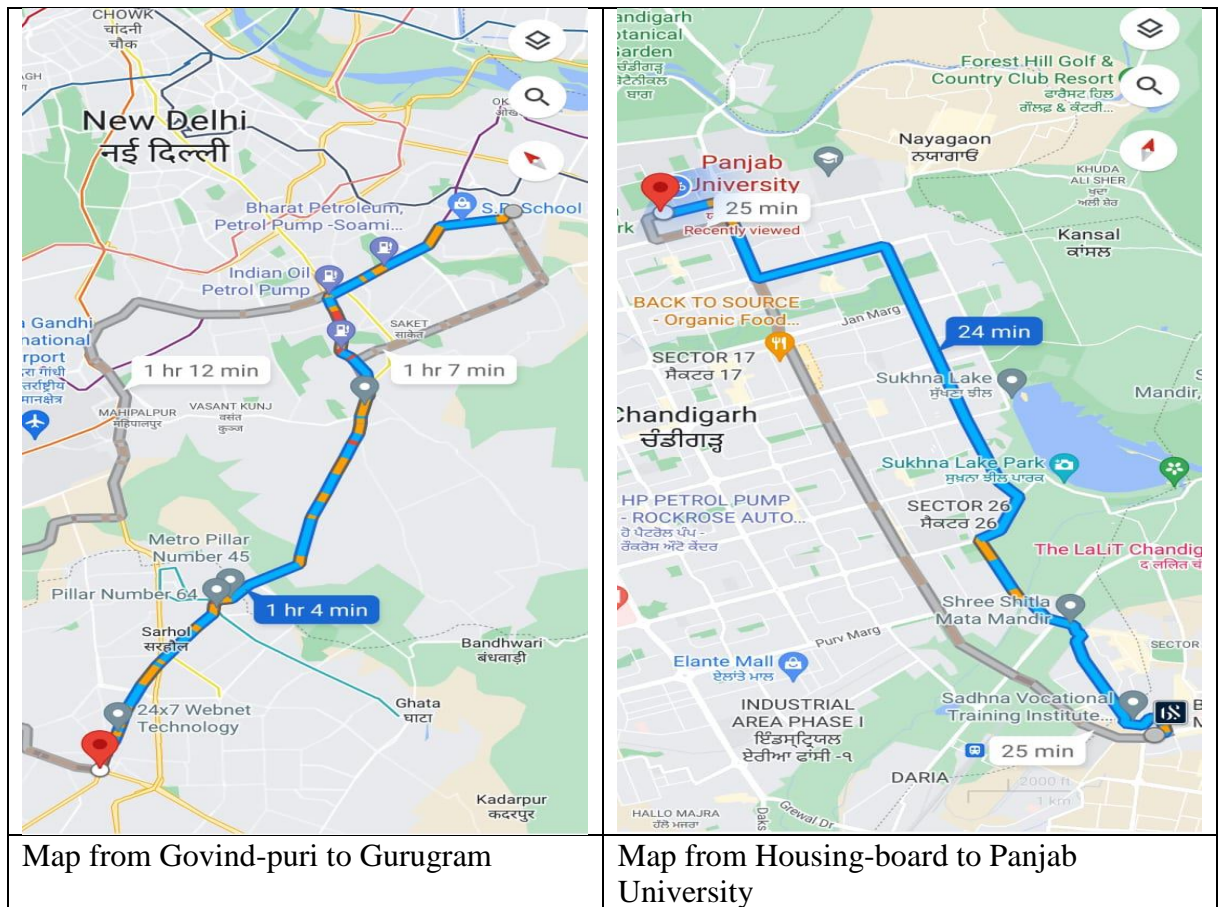

| Map from Govind-puri to Gurugram | Map from Housing-board to Panjab University |

**Figure 8:** Maps from Chandigarh and New Delhi Cities

**6. Video data:** In the process of creating our dataset, we first captured video sequences from the roads of New Delhi and Chandigarh. The main purpose of creating this dataset was to create a hybrid dataset having traffic lights from different cities having different orientations. Like in Chandigarh we have supported traffic lights and in New Delhi suspended traffic lights are there. After capturing video sequences, we extracted frames from the videos using the VLC media player. We followed the following steps for image extraction:

1. Open the VLC player and pause the video.
2. Then under the tools tab, we clicked on preferences.
3. After that002C we landed on the Interface tab, where we need to select "All" preferences under the "show settings" bar.
4. This will lead us to "Advanced Preferences", here under the video submenu we clicked on "filter" and select "scene filter" from the drop-down menu.
5. A new small window pane will open on the right side of the same dialogue box.

6. Here we can specify the image format, Height, width, filename prefix directory path, and the recording ratio.
7. The Recording ratio denotes the ratio of images to be recorded. Here recording ratio of 10 means one image out of ten images are stored. We kept recording ratio as 3.
8. Now after changing all the settings, we played the video. As soon as the video is done playing, go to the specified directory and here you can see images extracted from the video. Steps for frame extraction are given in the images from 1 to 4.

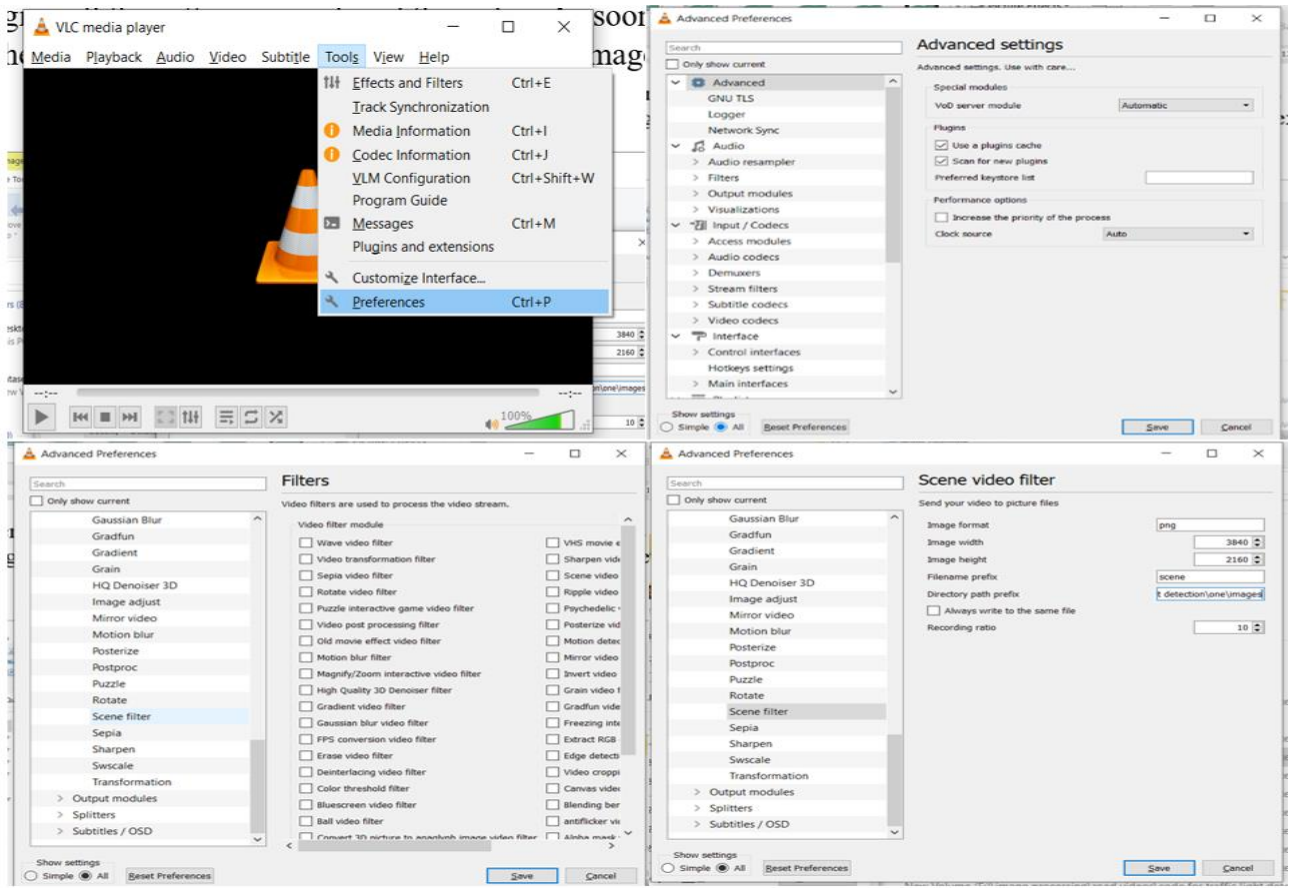

**Figure 9**: steps for image extraction from video sequences.

7. **Real-time images of the dataset in dayLight:** Images extracted from video frames in dayLight

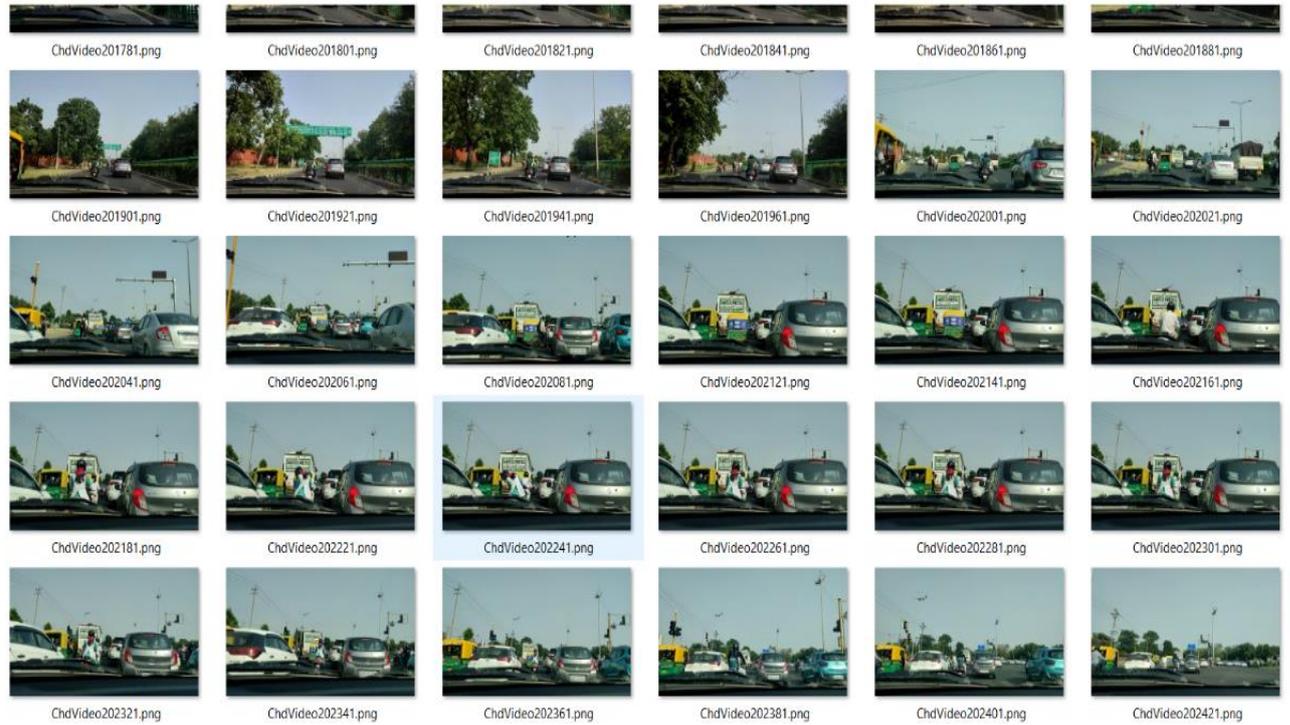

**Figure 10 :** Daylight images

## 8. Real-time images of the dataset in nightLight: Images extracted from video frames in nightLight

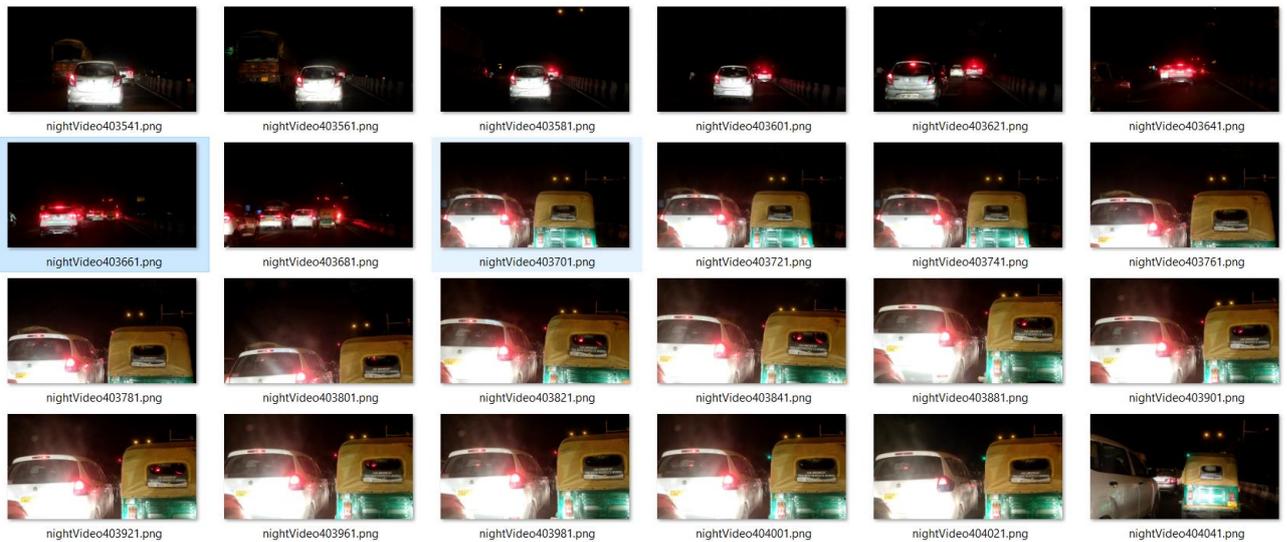

**Figure 11:** Nightlight images

## 9. Image Annotation:

For the purpose of labelling and annotation we used a freely available image annotation, tool "Super Annotate". In the labelling process, we first defined 14 classes with different colour codes namely car, traffic light, green traffic light, red-light, green-arrow-light, pedestrian green light, pedestrian-red-light, yellow light, timer, person, auto-rickshaw, bicycle, motor-bike, truck, bus, board as shown in Figure 3. We created squared shape bounding boxes and some free-hand boundaries for labelling the objects in the images. Our dataset contains a total of 3050 images with 3840 X 2160 resolution. From figure 5 to 7 various windows and dialogue boxes of Super-annotate is shown.

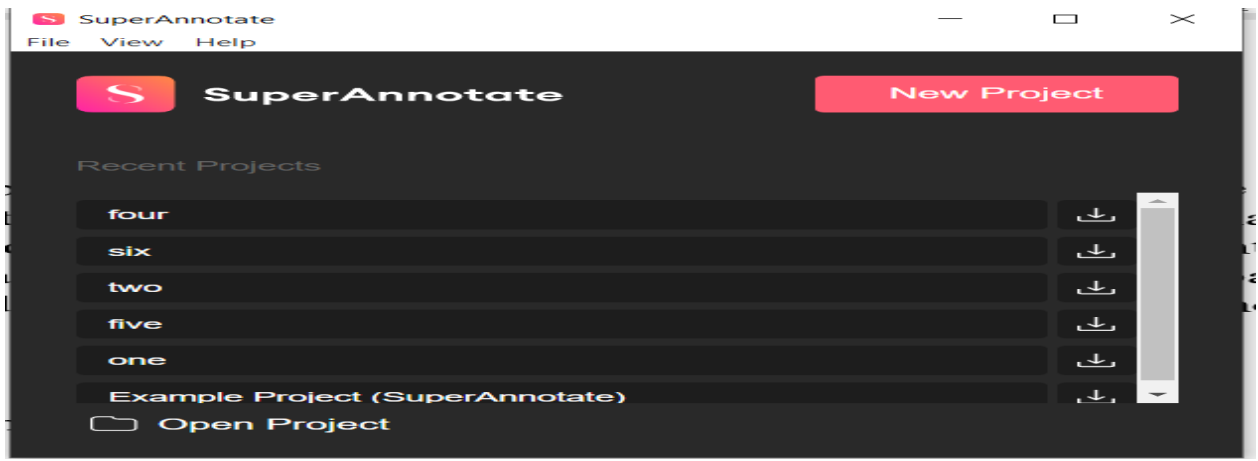

**Figure 12:** Super Annotate tool : for labelling images.

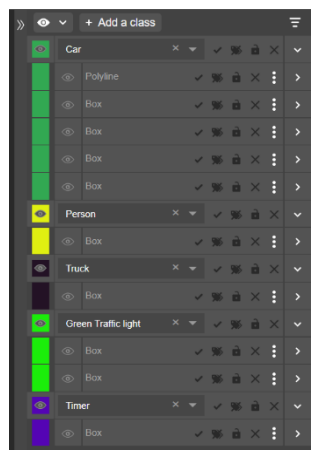

**Figure 13:** Classes with different colour codes for labelling images.

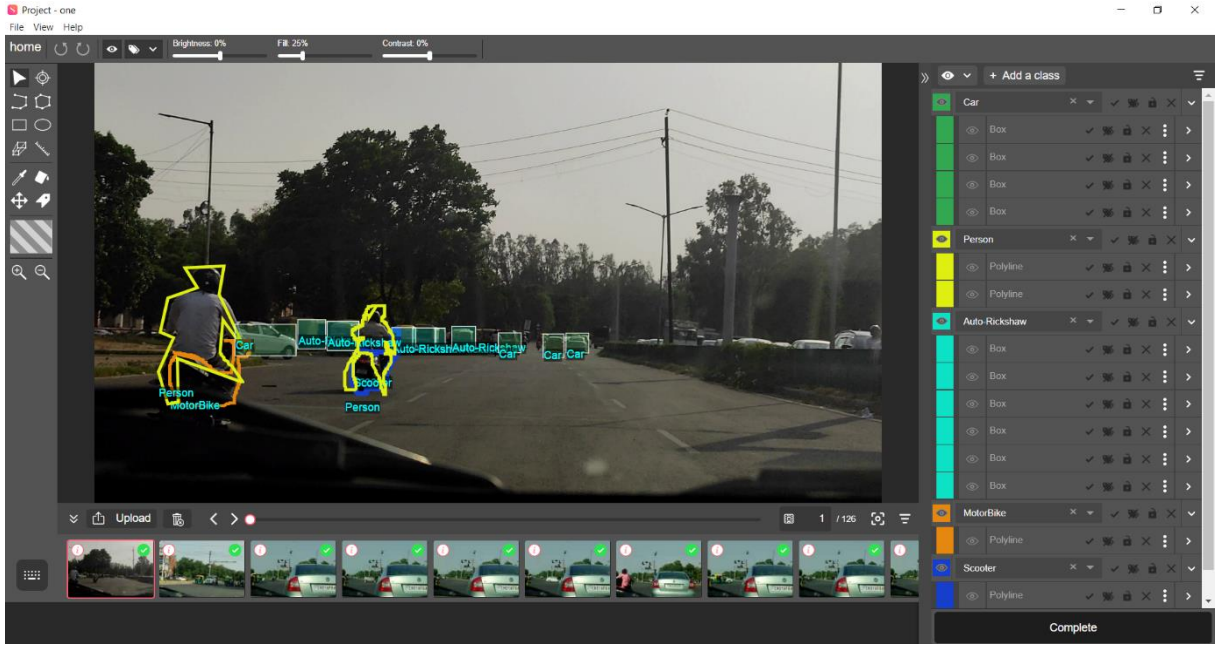

**Figure 14:** The labelled objects and bounding boxes

**10. Evaluation:** Transfer learning is a process of reusing pre-trained knowledge and applying it to process similar kinds of new data. During the evaluation process, we tried to detect traffic lights and some other objects like people, cars, and other vehicles. We used the single shot detector model for object detection and trained our Inception_v3 model on it. We created a traffic light extraction model to create cropped images of traffic lights and then detected traffic lights and other objects from the cropped images as shown in Figure 12. Here are some of the cropped images as shown in Table 5. We evaluated our dataset on an NVIDIA GEFORCE GTX 1650 ti having GB RAM. Our system achieved 97.23% test accuracy. The line graph is shown in Figure 13. The evaluation results are shown in Table 5 along with an infographics of extracted features in

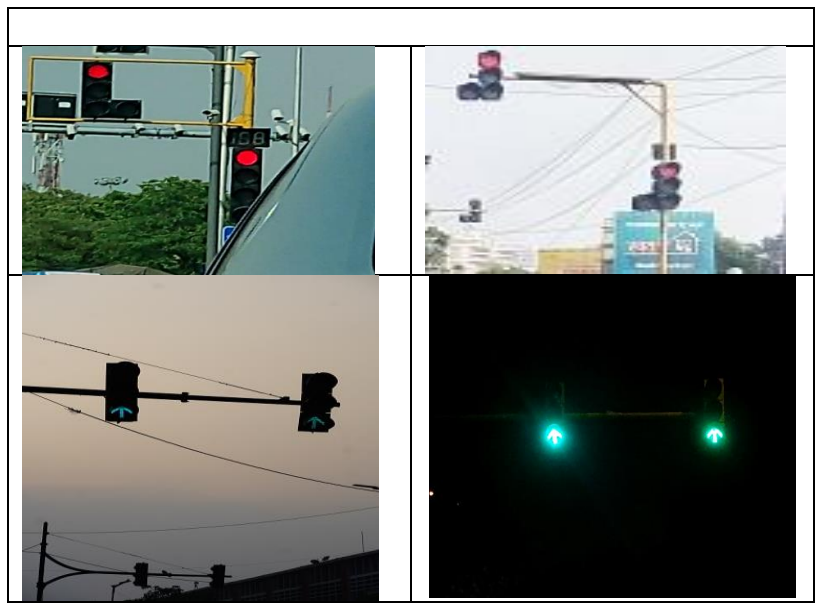

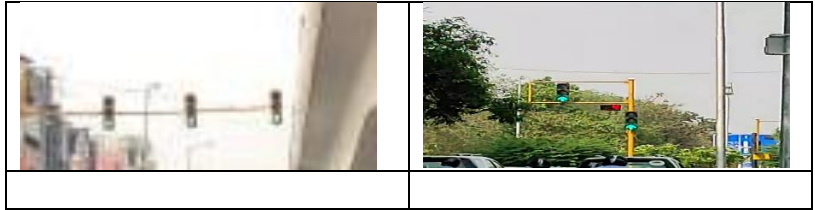

**Figure 15:** Cropped images of traffic lights using Single Shot Detector

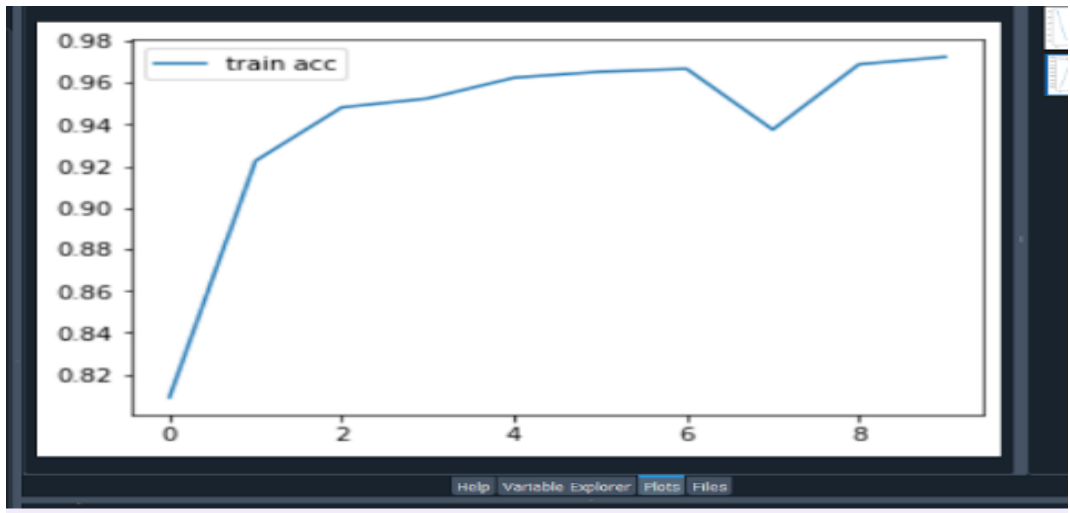

**Figure 16 :** Line graph of Detection accuracy using Inception_v3 model.

| S.no | Name of the folder | Number of images in each folder | Features extracted (Traffic light and similar looking objects) |
|---|---|---|---|
| 1. | One | 252 | 1496 |
| 2. | Two | 210 | 1331 |
| 3. | Three | 392 | 2187 |
| 4. | Four | 318 | 1053 |
| 5. | Five | 336 | 1343 |
| 6. | Six | 1139 | 780 |
| 7. | Seven | 353 | 1256 |
| 8. | Eight | 350 | 655 |
| 9. | Nine | 350 | 2507 |
| 10. | Ten | 274 | 160 |

**Table 5**: Shows number of features extracted from each folder.

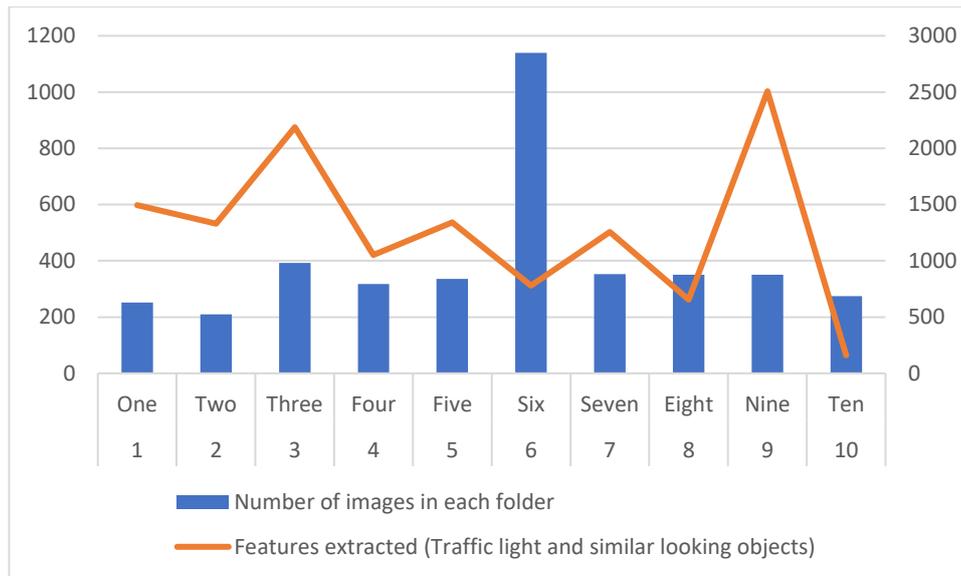

**Figure 17**: Infographic of extracted features

12. **Conclusion and Future scope:** A manually annotated labelled dataset IRD is presented in this paper which is based on two different cities having two different types of traffic light orientations. This dataset can be used to detect traffic lights and other objects like persons, cars, other vehicles, etc. This dataset provides insight into Indian roads. Which can help motor industrialists to gain better insight into Indian roads and manufacture autonomous cars accordingly. Here we applied transfer learning and Single shot detector methods to identify traffic lights in an image. The limitation of our dataset lies in the fact that there is a limited number of images in the night light and in the future we will extend our dataset to night light, foggy and rainy weather and automatically remove blurry images captured due to camera motion. We will try to test our dataset in real-time using a real vehicle.